\documentclass{article}

\usepackage{PRIMEarxiv}

\usepackage[utf8]{inputenc} % allow utf-8 input
\usepackage[T1]{fontenc}    % use 8-bit T1 fonts
\usepackage{hyperref}       % hyperlinks
\usepackage{url}            % simple URL typesetting
\usepackage{booktabs}       % professional-quality tables
\usepackage{amsfonts}       % blackboard math symbols
\usepackage{nicefrac}       % compact symbols for 1/2, etc.
\usepackage{microtype}      % microtypography
\usepackage{fancyhdr}       % header
\usepackage{graphicx}       % graphics
\usepackage{times}
\usepackage{epsfig}
\usepackage{graphicx}
\usepackage{subfigure}
\usepackage{multirow}
\usepackage{multicol}
\usepackage{cite}
\usepackage{amsmath,amssymb}
\usepackage{textcomp}
\usepackage{xcolor}
\usepackage{algorithm}  
\usepackage{algorithmicx}  
\usepackage{algpseudocode}
\graphicspath{{media/}}  

\title{Evolution: A Unified Formula for Feature Operators from a High-level Perspective
}

\author{
  Zhicheng Cai\\
 Nanjing University \\
  \texttt{181180002@smail.nju.edu.cn} \\
}

\begin{document}
\maketitle

\begin{abstract}
Traditionally, different types of feature operators (e.g., convolution, self-attention and involution) utilize different approaches to extract and aggregate the features. Resemblance can be hardly discovered from their mathematical formulas.
However, these three operators all serve the same paramount purpose and bear no difference in essence.
Hence we probe into the essence of various feature operators from a high-level perspective, transformed their components equivalently, and explored their mathematical expressions within higher dimensions.
We raise one clear and concrete unified formula for different feature operators termed as \textbf{\emph{Evolution}}. Evolution utilizes the \textbf{\emph{Evolution Function}} to generate the  \textbf{\emph{Evolution Kernel}}, which extracts and aggregates the features in certain positions of the input feature map. 
We mathematically deduce the equivalent transformation from the traditional formulas of these feature operators to Evolution and prove the unification.
In addition, we discuss the forms of Evolution Functions and the properties of generated Evolution Kernels, intending to give inspirations to the further research and innovations of powerful feature operators.
\end{abstract}

\section{Introduction}
Neural networks, represented by Convolutional Neural Network (CNN)~\cite{krizhevsky2012imagenet,simonyan2014very, he2016deep,huang2017densely}, Vision Multi-Layer Perceptrons (Vision MLP)~\cite{melas2021you,tolstikhin2021mlp,touvron2021resmlp} and Vision Transformers (ViT)~\cite{vaswani2017attention,liu2021swin,dosovitskiy2020image}, possess the capacity of representing complicated high-dimensional functions and process big data, which have successfully addressed many computer vision problems and implemented on real-world applications.
One vital element of these different visual neural networks (including these hybrid models) is utilizing specific feature operators to extract and aggregate the features in the feature maps. As a common practice,  CNNs utilize convolution (e.g., regular convolution, group convolution~\cite{krizhevsky2012imagenet}, depth-wise convolution~\cite{howard2017mobilenets} and point-wise convolution~\cite{howard2017mobilenets}) as the feature operator, MLPs utilize the feed-forward networks as the feature operator which can be exactly regarded as the combination of two point-wise convolutions, and ViTs utilize the self-attention. Moreover, there emerges many novel atom operators (e.g., involution~\cite{li2021involution,liang2021cnet}) which possess different characteristics and process the features in a different way.

Traditionally, convolution, self-attention~\cite{vaswani2017attention,liu2021swin,dosovitskiy2020image,ramachandran2019stand,ho2019axial} and involution~\cite{li2021involution,liang2021cnet} are regarded as three different types of feature operators, which utilize diverse approaches to process the information the input feature maps. From the perspective of their mathematical formulas simply, it fails to be discovered any resemblance superficially. However, these three operators all serve the same paramount purpose and bear no difference in essence. In addition, they all produce the output by aggregation local input features. As a result, they can not be viewed in a detached way. Moreover, we argue that it is required one clear and concrete formula to integrate and express these operators generally. 

As a matter of fact, there has emerged one simple general formula to describe basic image feature operators~\cite{hu2019local}. However, this formula only covers traditional and simple feature operators, such as standard convolution, deformable convolution~\cite{dai2017deformable}, and so on~\cite{yu2015multi,howard2017mobilenets,sandler2018mobilenetv2,chollet2017xception}. It fails to take more novel and complicated feature operators like self-attention and involution into consideration. As a result, it lacks the conclusion and proof that convolution, self-attention and involution can be unified in one formula. Moreover, the mathematical expression is extremely abstract and generalized, which is inconsistent with our original vision. 
While the other work~\cite{chen2021x,dai2021coatnet} tends to linearly combine convolution and self-attention with two learnable coefficients. It is a simple combination which ignores the mathematical essence of different feature operators, nevertheless figuring out a unified formula.  

Nevertheless, we probed into the essence of various feature operators from a high-level perspective, transformed their components equivalently, and explored their mathematical expressions within higher dimensions. 
We raised one mathematical architecture to express various deep learning feature operators, and we completed and proved the transformation from the traditional formulas of these operators to our unified formula. This paper will discuss the unified formula and the deduction of transformation in detail. 
In summary, we discovered one general formula for feature operators termed as \textbf{\emph{Evolution}}, which is both clear and concrete. Evolution utilizes the \textbf{\emph{Evolution Function}}, which takes the feature map as the input, to generate the  \textbf{\emph{Evolution Kernel}}, which extracts and aggregates the features in certain positions of the input feature map. In addition, we discuss the forms of Evolution Functions and the properties of generated Evolution Kernels. 
%Based on this unified formula, we raised one local feature operator family, termed as \textbf{\emph{Evolution}}. 
%We hope our unified formula can give inspirations to the further research and innovations of powerful deep learning feature operators.

\section{Traditional Formulas}

In this section, we recall the traditional formulas of three powerful feature operators, namely, convolution, self-attention and involution. We only discuss the operators essentially, ignoring the network architectures.

\subsection{Convolution Operator}

Firstly, we recall the formula of convolution layers. Given an input feature map tensor $\textbf{X} \in \mathbb{R}^{H_{in}\times W_{in}\times D_{in}}$ of height $H_{in}$, width $W_{in}$ and $D_{in}$ channels, the target of convolution is to learn a weight tensor $\textbf{W} \in \mathbb{R}^{K\times K\times D_{in}\times D_{out}}$ straightforwardly to aggregate local information and extract features. $\textbf{W}$ is shared across spatial position, which is regarded to be spatial-agnostic, enable translation invariance, and enhance the sparsity of parameters. As a consequence, we obtain the output feature map $\textbf{Y} \in \mathbb{R}^{H_{out} \times W_{out} \times D_{out}}$. Suspect the hyper-parameter stride $S$ and padding $P$, thus: 

\begin{equation}
H_{out} = \left\lfloor \frac{H_{in}+2P-K+S}{S} \right\rfloor,\  W_{out} = \left\lfloor\frac{W_{in}+2P-K+S}{S} \right\rfloor  
\label{eq1}
\end{equation}

For the convenience of expression, we consider the same convolution, discard the bias vector and set the group to be single. As a consequence, $H_{out} = H_{in} = H$, $W_{out} = W_{in}= =W$. The output feature map of a convolution layer is given by the formula below:

\begin{equation}
    \begin{split}
    &\textbf{Y}_{i,j,m} = Conv(\textbf{W},\textbf{X})_{i,j,m} = \sum_a \sum_b \sum_c \textbf{W}_{l-a,l-b,c,m}\textbf{X}_{i-a,j-b,c}\\
    \label{eq2}
    &\textbf{where}\ \ a,b \in [-l,l]\ , \  l = \left\lfloor\frac{K}{2}\right\rfloor \ , \ c \in [0,D_{in}-1] 
    \end{split}
\end{equation}

where $a, b$ stand for the possible vertical shift and horizontal shift respectively when convolving the input feature map with a $K\times K$ convolution kernel.

When considering $y_{ij} \in \mathbb{R}^{D_{out}}$ which stands for the pixel $(i,j)$ in feature map $Y$, we obtain:

\begin{equation}
    \begin{split}
    \textbf{y}_{ij} &= \sum_a \sum_b \sum_c \textbf{W}_{l-a,l-b,c,:}\textbf{X}_{i-a,j-b,c} \\
    &= \sum_a \sum_b \textbf{W}_{l-a,l-b,:,:}\textbf{X}_{i-a,j-b,:} \\
    \label{eq3}
    \end{split}
\end{equation}

\subsection{Self-Attention Operator}

Here we recall the mathematical formula of self-attention layers. 

\subsubsection{Original Self-Attention}

As NLP is the cradle for self-attention mechanism, recalling self-attention in NLP realm is supposed to be meaningful. Given $\textbf{X} \in \mathbb{R}^{N\times D_{in}}$ which is an input matrix consisting of $N$ features with $D_{in}$ dimensions each. Features here can refer to any sequence of $N$ discrete samples, such as tokens and pixels. The formula of one self-attention layer mapping the input matrix $\textbf{X}$ into output $\textbf{Y} \in \mathbb{R}^{N\times D_{out}}$ can be written as below:

\begin{equation}
    \begin{split}
    &\textbf{Y}_{n,m} = \sigma(\textbf{A}_{n,:}){(\textbf{X}\textbf{W}_V)}_{:,m}\\
    \\
    &\textbf{where} \ \ \textbf{A} = \textbf{X}\textbf{W}_Q\textbf{W}^T_K\textbf{X}^T \\
    \\
    &\sigma(\textbf{A}_{n,:})_i = \frac{exp(\textbf{A}_{n,i})}{\sum_k exp(\textbf{A}_{n,k})}
    \label{eq4}
    \end{split}
\end{equation}

In this formula, $\textbf{W}_Q \in \mathbb{R}^{D_{in}\times D_{k}}, \textbf{W}_K \in \mathbb{R}^{D_{in}\times D_{k}}, \textbf{W}_V \in \mathbb{R}^{D_{in}\times D_{out}}$ stand for the query matrix, key matrix and value matrix correspondingly. These three matrices are the learned weights in each self-attention layer. Matrix $\textbf{A} \in \mathbb{R}^{D_{in}\times D_{in}} $ represents the map of attention scores. Function $\sigma(\textbf{A})$ represents the softmax function, which produces the map of attention probabilities. 

Accordingly, for $y_{n}$ represents the n-th sample in map Y, we obtain: 
\begin{equation}
\textbf{y}_{n} = \sigma(\textbf{A}_{n,:}){(\textbf{X}\textbf{W}_V)}
\label{eq5}
\end{equation}

\subsubsection{Self-Attention with Positional Encoding}

The self-attention mechanism discussed above possesses a critical characteristic that the connection distance to each sample is equal, which means it is equivalent to reordering essentially and produces the same output disregarding how the input $N$ samples shuffled. This can be confused when the order of samples matters. To alleviate this issue, positional encoding is raised, which can be learned or substituted by any function that produce vectors contain the representation of position. Suppose $P\in \mathbb{R}^{N\times D_{in}}$ contains the embedding vectors for each position, $P$ is added to the representation of the samples before conducting self-attention, thus we obtain the attention scores map $\textbf{A}_P$:

\begin{equation}
\textbf{A}_P = (\textbf{X+P})\textbf{W}_Q \textbf{W}^T_K(\textbf{X+P})^T 
\label{eq6}
\end{equation}

\subsubsection{Multi-Head Self-Attention}

It is found to be beneficial to split the self-attention into multiple heads, each learning different query, key and value matrices to extract different features. The projections of $M$ heads are performed simultaneously, the outputs $\in \mathbb{R}^{N \times D_h}$ ($D_h = D_{out}/M$ for the most cases) are concatenated and projected by $\textbf{W} \in \mathbb{R}^{MD_h \times D_{out}}$ to dimension $D_{out}$:

\begin{equation}
    \begin{split}
    &\textbf{Y} = concat(\textbf{SA}_0, \cdot\cdot\cdot, \textbf{SA}_{M-1})\textbf{W} \\
    &\textbf{SA}_i = \sigma(\textbf{X} {\textbf{W}_{Q}}_{i} {\textbf{W}^T_{K}}_{i} \textbf{X}^T)\textbf{X}{\textbf{W}_{V}}_{i} \\
    &\textbf{where} \ \ {\textbf{W}_{Q}}_{i}, {\textbf{W}_{K}}_{i}, {\textbf{W}_{V}}_{i} \in 
    \mathbb{R}^{D_{in}\times D_{h}}
    \label{eq7}
    \end{split}
\end{equation}

\subsubsection{Self-Attention for Image}

Now we consider the circumstance of images. Given an input feature map $\textbf{X} \in \mathbb{R}^{H\times W\times D_{in}}$, suppose that the size of feature map is constant, we want to obtain the output feature map $\textbf{Y} \in \mathbb{R}^{H\times W\times D_{out}}$ by applying self-attention. It can be expressed as:

\begin{equation}
    \begin{split}
    &\textbf{Y}_{i,j,m} = SA(\textbf{W},\textbf{X})_{i,j,m} =\sum_a \sum_b \sigma_{ab}(\textbf{X}_{i,j,:}\textbf{W}_Q\textbf{W}^T_K \textbf{X}^T_{i-a,j-b,:})(\textbf{X}_{i-a,j-b,:}\textbf{W}_V)_{:,m}\\
    \label{eq8}
    &\textbf{where}\ \ a \in [i-H,i]\ ,\  b \in [j-W,j]
    \end{split}
\end{equation}

Considering the computationally expensive, global attention can only be utilized after the application of significant downsampling to the input image, thus we probe to the local single-headed self-attention. Given the size of aggregation area is $K\times K$, The pixel output $y_{ij} \in \mathbb{R}^{D_{out}}$ can be obtained as follows:

\begin{equation}
    \begin{split}
    &\textbf{y}_{ij} =\sum_a \sum_b \sigma_{ab}(\textbf{X}_{i,j,:}\textbf{W}_Q\textbf{W}^T_K \textbf{X}^T_{i-a,j-b,:})(\textbf{X}_{i-a,j-b,:}\textbf{W}_V)\\
    \label{eq9}
    &\textbf{where}\ \ a,b \in [-l,l]\ , \  l = \left\lfloor\frac{K}{2}\right\rfloor
    \end{split}
\end{equation}

where $\sigma_{ab}(x)$ conducts the softmax operation on all the neurons in the neighborhood of $ij$. In this paper, we discuss about local self-attention mainly. The cases of positional encoding and multi-head self-attention is similar to Eq.~\ref{eq6} and Eq.~\ref{eq7} respectively.

\subsection{Involution Operator}

Involution is an operator which has been proposed recently. Contrast to the inherence of standard convolution that is spatial-agnostic and channel-specific, involution possesses the property of spatial-specific and channel-agnostic. Similarly, let $\textbf{X} \in \mathbb{R}^{H\times W\times D}$ be the input tensor, the output tensor obtained is $\textbf{Y} \in \mathbb{R}^{H\times W\times D}$, possessing the same size with $\textbf{X}$. The channels number of input $\textbf{X}$ and output $\textbf{Y}$ are the same.  Considering involution kernel to be $\textbf{W} \in \mathbb{R}^{H\times W\times K \times K\times G}$, the output feature map produced can be formulated as:

\begin{equation}
    \begin{split}
    &\textbf{Y}_{i,j,m} = Inv(\textbf{W},\textbf{X})_{i,j,m} =\sum_a \sum_b \textbf{W}_{i,j,l-a,l-b,\lceil kG/D\rceil} \textbf{X}_{i-a,j-a,m} \\
    &\textbf{where}\ \ a,b \in [-l,l]\ , \  l = \left\lfloor\frac{K}{2}\right\rfloor
    \label{eq10}
    \end{split}
\end{equation}

where the involution kernel $\textbf{W}$ is comfortably aligned to the input feature map $\textbf{X}$. As a consequence, one involution kernel generation function $\phi$ is required. Given $\textbf{X}_{\Psi_{i,j}}$, which is one local patch of $\textbf{X}$ covering a scope of $\Psi_{i,j} \in \mathbb{R}^{a\times b\times D}$, as the prerequisite, the involution kernel $\textbf{W}_{i,j}$ produced at location $(i,j)$  can be formulated as:

\begin{equation}
\textbf{W}_{i,j} = \phi(\textbf{X}_{\Psi_{i,j}})
\label{eq11}
\end{equation}

Generation function $\phi$ possesses various potential formulas, paper involution utilized a relatively simple function $\phi$ to generate the involution kernel, of which the hyper-parameter $G$ is set to be one. In addition, the local patch $\textbf{X}_{\Psi_{i,j}}$ conditioned on is parochial, covering singleton pixel $\textbf{X}_{i,j} \in \mathbb{R}^{D}$ only. Formally, the projection of function $\phi$ is expressed as:

\begin{equation}
\phi(\textbf{X}_{i,j}) = \textbf{W}_1\sigma(\textbf{W}_0\textbf{X}_{i,j})
\label{eq12}
\end{equation}

where $\textbf{W}_0 \in \mathbb{R}^{\frac{D}{r}\times D}$ and $\textbf{W}_1 \in \mathbb{R}^{(K^2)\times \frac{D}{r}}$ stand for two linear projections composing a bottleneck architecture. The hyper-parameter $r$ implies the channel dimension attenuation ratio for efficiency. Besides, function $\sigma$ represents batch normalization cascaded with one non-linear activation function. The output $\phi(\textbf{X}_{i,j})$ possesses the size of $1\times 1\times K^2$, then it is resized to the shape of $K\times K\times 1$ to obtain the final involution kernel $\textbf{W}_{i,j} \in \mathbb{R}^{K\times K \times 1}$. This procedure is believed to obtain a channel-to-space transformation. During realistic computation, $\textbf{W}_{i,j}$ is broadcast to $D$ channels, thus we obtain $\textbf{W}_{i,j} \in \mathbb{R}^{K\times K \times D}$, sharing the same values in the third dimension. Then it performs multiplication and aggregation sequentially. The pixel $y_{ij}$ of the corresponding output feature map $Y$ can be formulated as below:

\begin{equation}
    \begin{split}
    &\textbf{y}_{ij} =\sum_a \sum_b \textbf{W}_{i,j,l-a,l-b,:} \textbf{X}_{i-a,j-a,:} \\
    &\textbf{where}\ \ a,b \in [-l,l]\ , \  l = \left\lfloor\frac{K}{2}\right\rfloor
    \label{eq13}
    \end{split}
\end{equation}

Essentially, the procedure of kernel generation depends on the information of channels specifically, especially when the local patches conditioned on are consist of one single pixel. Besides, the kernel generation method shares weights in all pixels. It is believed that involution is channel-agnostic and spacial-specific for that each position $(i,j)$ in the input feature map shares one exclusive aggregation kernel, nevertheless, we argue that involution is still one spatial-agnostic and channel-specific method from a meta perspective. Moreover, during the procedure of weight aggregation, the involution operation ignores the information across different channels by taking a \emph{depth-wise} like aggregation approach.

\section{Evolution: A Unified Formula}

In this section, we will first give the unified formula termed as \textbf{\emph{Evolution}}. Then, we will infer the transformation of three typical operators from their traditional formulas to our unified formula in detail. 

\subsection{The Unified Formula}
Here we express the unified formula first. Similarly, given input feature map $\textbf{X} \in \mathbb{R}^{H_{in}\times W_{in}\times D_{in}}$, the output feature map produced is $\textbf{Y} \in \mathbb{R}^{H_{out}\times W_{out}\times D_{out}}$. In addition, one aggregation weight tensor $\textbf{W} \in \mathbb{R}^{H_{in}\times W_{in}\times K\times K \times N \times D_{out}}$ is required, where $K\times K$ implies the local aggregation scope of the input feature map, $N = (D_{in}/G)$ stands for the aggregation scope in the channel dimension, and $G$ represents the parameter of group. This aggregation weight tensor is termed as \textbf{Evolution Kernel}, which can still be regarded as one sliding window on the feature map except for that its values can be changeable and specific. The evolution kernel can be obtained by a certain kernel generation function $\textbf{F}$ termed as \textbf{Evolution Function}.  For the convenience of statement, we suppose that $H_{in} = H_{out} = H$, $W_{in} = W_{out} = H$. Thus, the unified formula can be expressed as below:

\begin{equation}
    \begin{split}
    &\textbf{Y}_{i,j,m} = Ev(\textbf{W},\textbf{X})_{i,j,m} = \sum_a \sum_b \sum_c \textbf{X}_{i-a,j-b,c}\textbf{W}_{i,j,l-a,l-b,c-N\times m,m}\\
    &\textbf{W} = \textbf{F}(\textbf{X}) \\
    &\textbf{where}\ \ a,b \in [-l,l]\ , \  l = \left\lfloor\frac{K}{2}\right\rfloor \ , \ c \in [N\times m,N\times(m+1)-1] 
    \label{eq14}
    \end{split}
\end{equation}

\subsection{Case of Convolution}
In respect of convolution, the kernel $\textbf{W}_{conv}$ possesses a size of $K\times K\times D_{in} \times D_{out}$ (suppose one group), which is learned directly and spatially-shared. Compared to the expression of evolution kernel, $\textbf{W}_{conv}$ lacks two dimensions, which means $\textbf{W}_{conv}$ fails to considering the possible variety spatially. To put it into the unified formula, we just need to the enhance the dimension of the original convolution kernel and duplicate $\textbf{W}_{conv}$ in each position $(i,j)$ of $\textbf{W}$, that is:

\begin{equation}
    \begin{split}
    &\textbf{Y}_{i,j,m} = \sum_a \sum_b \sum_c \textbf{X}_{i-a,j-b,c}\textbf{W}_{i,j,l-a,l-b,c,m}\\
    &\textbf{where}\ \ \textbf{W}_{i,j} = \textbf{W}_{conv} \\
    \label{eq15}
    \end{split}
\end{equation}

From the perspective of evolution kernel generation, there still exists one function to generate convolution kernels. This function $F$ actually accepts identity projection tensor $\textbf{I} \in \mathbb{R}^{K\times K\times D_{in} \times D_{out}}$ as the input. That can be expressed as:

\begin{equation}
    \begin{split}
    \textbf{W}_{i,j} = F_{i,j} = F(I)_{i,j} = \textbf{I}\textbf{W}_{conv} \\
    %\textbf{where}\ \ i \in [0,H&-1]\ , \  j \in [0,W-1]
    \label{eq16}
    \end{split}
\end{equation}

\subsection{Case of Self-attention}

\subsubsection{Evolution version of self-attention }

As for the aspect of self-attention, we can regard it to be a kernel generation function from a high-level perspective. It is established that matrices $\textbf{W}_Q \in \mathbb{R}^{D_{in}\times D_{k}}, \textbf{W}_K \in \mathbb{R}^{D_{in}\times D_{k}}, \textbf{W}_V \in \mathbb{R}^{D_{in}\times D_{out}}$ are learned parameters. If we add two dimensions to these three matrices, which become $\textbf{W}_Q \in \mathbb{R}^{1\times 1\times D_{in}\times D_{k}}, \textbf{W}_K \in \mathbb{R}^{1\times 1\times D_{in}\times D_{k}}, \textbf{W}_V \in \mathbb{R}^{1\times 1\times D_{in}\times D_{out}}$ respectively. Thus We can regard these three tensors as convolution kernels with a kernel size of $1\times 1$. As a consequence, we can transform the procedure of getting  queries and keys from original matrix multiplication into standard convolution as expressed below:

\begin{equation}
    \begin{split}
    &\textbf{Q}_{i,j,m} = \textbf{X}_{i,j,:}{\textbf{W}_Q}_{:,m} = Conv(\textbf{W}_Q,\textbf{X})_{i,j,m}\\
    &\textbf{K}_{i,j,m} = \textbf{X}_{i,j,:}{\textbf{W}_K}_{:,m} = Conv(\textbf{W}_K,\textbf{X})_{i,j,m} \\
    &\textbf{V}_{i,j,m} = \textbf{X}_{i,j,:}{\textbf{W}_V}_{:,m} = Conv(\textbf{W}_V,\textbf{X})_{i,j,m} \\
    \label{eq17}
    \end{split}
\end{equation}

Now we have obtained the query tensor $\textbf{Q} \in \mathbb{R}^{H\times W\times D_{k}}$, the key tensor $\textbf{K} \in \mathbb{R}^{H\times W\times D_{k}}$, and the value tensor $\textbf{V} \in \mathbb{R}^{H\times W\times D_{out}}$. Similarly, we can utilize convolution to transform the procedure of acquiring the attention scores map $\textbf{A} \in \mathbb{R}^{H\times W\times K\times K}$. Suppose that $\textbf{K}_{i,j,ab}$ covers one $K\times K$ local patch with the center of pixel $(i,j)$, we create a new key tensor $\textbf{K}^{\prime} \in \mathbb{R}^{H\times W\times K\times K \times D_{k}}$, where $\textbf{K}^{\prime}_{i,j} = \textbf{K}_{i,j,ab}$. Besides, we increase the dimension of $\textbf{Q}$, obtaining $\textbf{Q}^{\prime} \in \mathbb{R}^{H\times W\times 1\times 1\times D_{k}\times 1}$. Consequently, $\textbf{Q}^{\prime}_{i,j}$ can be regarded one $1\times 1$ convolution kernel, with $D_{k}$ input channels and one output channel. Moreover, $\textbf{K}^{\prime}_{i,j}$ can be considered as one input feature map of height $K$, width $K$ and $D_{k}$ channels. Additionally, the procedure of obtaining attention score map $\textbf{A}$ can be written in the evolution formula as well. Actually, $\textbf{Q}^{\prime}$ and $\textbf{K}^{\prime}$ can be regarded as the evolution kernel and the input feature map respectively. That is:

\begin{equation}
    \begin{split}
    \textbf{A}_{i,j,m,n} &= \textbf{X}_{i,j,:}\textbf{W}_Q\textbf{W}^T_K\textbf{X}^T_{i-l+m,j-l+n,:} \\
                         &= Conv(\textbf{Q}^{\prime}_{i,j},\textbf{K}^{\prime}_{i,j})_{m,n} \\
                         &= Ev(\textbf{Q}^{\prime},\textbf{K}^{\prime}_{:,:,m,n,:})_{i,j} \\
    \textbf{where}\ \  &m,n \in [0,K]\ , \  l = \left\lfloor\frac{K}{2}\right\rfloor \
    \label{eq18}
    \end{split}
\end{equation}

At present, if we calculate the attention probability $\sigma_{ab}(\textbf{A})$ with $\textbf{V}$, according to Eq.~\ref{eq8} we obtain:

\begin{equation}
    \begin{split}
    &\textbf{Y}_{i,j,m} = \sum_a \sum_b \textbf{V}_{i-a,j-b,m}{\sigma_{ab}(\textbf{A}_{i,j})}_{l-a,l-b}\\
    &\textbf{where}\ \ a,b \in [-l,l]\ , \  l = \left\lfloor\frac{K}{2}\right\rfloor 
    \label{eq19}
    \end{split}
\end{equation}
 
Compare Eq.~\ref{eq19} with Eq.~\ref{eq14}, we can find that Eq.~\ref{eq19} firstly perform the projection to the original input feature map, while the input feature map is unchanged before computing with the evolution kernel in Eq.~\ref{eq14}. In addition, the kernel $\textbf{A}$ lacks two dimensions compared to the standard evolution kernel. Actually, the computation between $\textbf{A}$ with values can still be accomplished by evolution operation. We only need to add two dimensions to $\textbf{A}$, that is, $\textbf{A}^{\prime}_{i,j} \in \mathbb{R}^{K\times K\times 1\times 1}$, then Eq.~\ref{eq19} becomes:

\begin{equation}
    \begin{split}
    \textbf{Y}_{i,j,m} &= \sum_a \sum_b ({\sigma_{ab}(\textbf{A}^{\prime}_{i,j})},\textbf{V}_{i-a,j-b,m})\\
    & = Conv({\sigma_{ab}(\textbf{A}^{\prime}_{i,j})},\textbf{V}_{i-a,j-b,m}) \\
    & = Ev(\sigma_{ab}(\textbf{A}^{\prime}),\textbf{V})_{i,j,m} \\
    & = Ev(\sigma_{ab}(\textbf{A}^{\prime}),Conv(\textbf{W}_{V},X))_{i,j,m} \\
    \textbf{where}&\ \  a,b \in [-l,l]\ , \  l = \left\lfloor\frac{K}{2}\right\rfloor 
    \label{eq20}
    \end{split}
\end{equation}

From the perspective of Eq.~\ref{eq20}, we can regard self-attention as two cascade evolution layers with no nonlinear activation functions intermediately. However, it still bears difference with our unified formula Eq.~\ref{eq14}. The key point to bridge the gap between Eq.~\ref{eq20} and Eq.~\ref{eq14} lies in the establish of the hypothesis below:

\begin{equation}
    \begin{split}
    Ev(\sigma_{ab}(\textbf{A}^{\prime}),Conv(\textbf{W}_{V},X))_{i,j,m} = Ev(Conv(\textbf{W}_{V},\sigma_{ab}(\textbf{A}^{\prime})),X)_{i,j,m}
    \label{eq21}
    \end{split}
\end{equation}

Now the problem becomes the proof of Eq.~\ref{eq21}. It is confused that whether there exists one weight tensor $\textbf{W}_{V}$ that the projection it performs one the input feature map can be absorbed into the $\textbf{A}^{\prime}$. Thus the production of the transformation is the final evolution kernel. We declare that the transformation expressed by Eq.~\ref{eq21} is practical. As exhibited above, the multiplication performed by the value matrix can be viewed as a $1\times 1$ convolution layer. $1\times 1$ convolution is known as the \emph{point-wise} convolution which is utilized to aggregate the information cross different channels. As a result, the essential effect of value matrix is working in the dimension of the channel. For $\textbf{X}_{i,j,m}$ in the pixel $(i,j)$, they share the same attention probability $\sigma_{ab}(\textbf{A}_{i,j})$, which means the attention probability map makes no difference to the information cross the channels. At present, out prime mission becomes how to make the attention probability map contains the transformation of channels. For each attention probability patch $\sigma_{ab}(\textbf{A}_{i,j})$, we expand it to $\textbf{A}^{\prime} \in \mathbb{R}^{K\times K\times D_{in}\times D_{out}}$, where $\textbf{A}^{\prime}_{:,:,c,m} = \sigma_{ab}(\textbf{A}_{i,j})$. Then for each element in $\textbf{A}^{\prime}_{:,:,c,m}$, we multiply it with ${\textbf{W}_V}_{:,:,c,m}$. That is:

\begin{equation}
    \begin{split}
    \textbf{A}^{\prime}_{:,:,c,m} = Conv({\textbf{W}_{V}}_{:,:,c,m},\sigma_{ab}(\textbf{A}_{i,j})) \\
    \label{eq22}
    \end{split}
\end{equation}

$\textbf{A}^{\prime}_{:,:,:,n}$ can be utilized to aggregate the information cross the channel. In addition, the parameter $n$ is exploited to generate the n-th output channel specifically, which is $D_{out}$ in total. Moreover, the total number of learned parameter remains the same after the transformation. Actually, $\textbf{A}^{\prime}$ is specific for each $\textbf{X}_{i,j}$. If we add the dimensions of position to $\textbf{A}^{\prime}$, it will be the prototype of the evolution kernel for self-attention.

Actually, the operation performed on each pixel $(i,j)$ can be regarded as \emph{depth-wise} convolution, except that the kernel is spatial-specific. Nevertheless, it can be regarded as \emph{depth-wise} evolution. Same to standard convolution, one standard evolution can be departed into one \emph{depth-wise} evolution and one \emph{point-wise} evolution. Considering there exists no activation functions intermediately, we can obtain the evolution kernel comprised according to the \emph{depth-wise} and \emph{point-wise} evolution kernels.

In summary, we claim that the self-attention in our unified formula can be written as:

\begin{equation}
    \begin{split}
    &\textbf{Y}_{i,j,m} = \sum_a \sum_b \sum_c \textbf{X}_{i-a,j-b,c} Conv({\textbf{W}_{V}}_{:,:,c,m},\sigma_{ab}(\textbf{A}_{i,j}))_{l-a,l-b} \\
    &\textbf{where}\ \ a,b \in [-l,l]\ , \  l = \left\lfloor\frac{K}{2}\right\rfloor \ , \ c \in [0,D_{in}-1] 
    \label{eq23}
    \end{split}
\end{equation}

From the perspective of evolution kernel generation, the evolution kernel generation function $F$ actually accepts the input tensor $\textbf{X}$ as the input. It is obviously that the evolution kernel inferred in the self-attention unified formula can be expressed as:

\begin{equation}
    \begin{split}
    &\textbf{W}_{i,j,a,b,c,m} = F(\textbf{X})_{i,j,a,b,c,m} = Conv({\textbf{W}_{V}}_{:,:,c,m},\sigma_{ab}(\textbf{A}_{i,j}))_{a,b} \\
    \label{eq24}
    \end{split}
\end{equation}

\subsubsection{Considering positional encoding}

If the positional encoding is required in the self-attention, the unified formula is still suitable.  As shown in Eq.~\ref{eq6}, the positional encoding only alternates the tensor $\textbf{Q}$ and $\textbf{K}$, making no difference to the tensor $\textbf{V}$. Consequently, the attention score map $\textbf{A}$ is alternated and computed accordingly, while the input feature map $\textbf{X}$ remains unchanged in Eq.~\ref{eq23}.  

\subsubsection{Considering multiple heads}

Here we discuss the case that the multi-head self-attention mechanism is applied to the feature operator. According to Eq.~\ref{eq7} where the number of heads is $M$, we split $\textbf{W}_{Q}$, $\textbf{W}_{K}$, and $\textbf{W}_{V} \in \mathbb{R}^{1\times 1\times D_{in} \times D_{k}}$ into ${\textbf{W}_{Q}}^{p}$, ${\textbf{W}_{K}}^{p}$, and ${\textbf{W}_{V}}^{p} \in \mathbb{R}^{1\times 1\times D_{in} \times D_{k}/M}, p\in [1,M]$. For each ${\textbf{W}_{Q}}^{p}$, ${\textbf{W}_{K}}^{p}$, and ${\textbf{W}_{V}}^{p}$, we obtain $\textbf{Q}^{p}$, $\textbf{K}^{p}$, and $\textbf{V}^{p}$ according to Eq.~\ref{eq17}. Afterwards we calculate the attention scores map $\textbf{A}^{p} \in \mathbb{R}^{H\times W\times K\times K}$ according to Eq.~\ref{eq18} and the output $\textbf{Y}^{p} \in \mathbb{R}^{H\times W\times D_{h}}$ according to Eq.~\ref{eq23}. Then we concatenate $\textbf{Y}^{0},\cdot\cdot\cdot,\textbf{Y}^{M-1}$ and project them with $\textbf{W}_{O}$ to obtain the final output. Similarly to Eq.~\ref{eq7}, the procedure can be expressed as:

\begin{equation}
    \begin{split}
    &\textbf{Y} = concat(\textbf{Y}^0,\cdot\cdot\cdot,\textbf{Y}^{M-1})\textbf{W}_{O} \\
    &\textbf{where} \ \ \textbf{Y}^{p}_{i,j,m} = Ev(\textbf{W}^{p},\textbf{X})_{i,j,m}\\ 
    &\textbf{W}^{p}_{i,j,a,b,c,m} = {Conv({\textbf{W}^{p}_{V}}_{:,:,c,m},\sigma_{ab}(\textbf{A}^{p}_{i,j}))}_{l-a,l-b} \\
    &\textbf{A}^{p}_{i,j,m,n} = Ev({\textbf{Q}^{\prime}}^{p},{\textbf{K}^{\prime}}^{p}_{:,:,m,n,:})_{i,j}
    \label{eq25}
    \end{split}
\end{equation}

It is obvious that Eq.~\ref{eq25} is not included in the unified formula, which is not consistency with out vision. So here is the problem, whether can we express the multi-head self-attention in the form of evolution? As a matter of fact, the answer is positive. The key point to accomplish this proof lies in the generation of the evolution kernel. Let us focus on the Eq.~\ref{eq19} firstly, the attention score is shared with each element in the dimension of channel $\textbf{W}_{i,j,:}$ in pixel $(i,j)$. However, for multi-head self-attention, there are multiple attention score maps, the n-th attention score map is shared in the channel range of $[(n-1)\times D_{h},n\times D_{h}]$. That's to say, the evolution kernel $\textbf{W} \in \mathbb{R}^{H\times W\times K\times K \times D_{in}\times D_{out}}$ in Eq.~\ref{eq23} is derived from the same attention score map across the channel dimension, one the other hand, the evolution kernel $\textbf{W}^{p} \in \mathbb{R}^{H\times W\times K\times K \times D_{in}\times D_{h}}$ is derived from difference attention score map, respectively corresponding to the specific channel interval. Consequently, we create a new evolution kernel $\textbf{W}^{\prime}$ by concatenating $\textbf{W}^{p}$ in the channel dimension. Thus we solve the first problem of concatenating the various outputs in Eq.~\ref{eq25}. As for the remaining linear projection $\textbf{W}_{O}$, in the light of the discussion above, we can squeeze it into $\textbf{W}^{\prime}$ correspondingly. Now we obtain the final evolution kernel as shown in Eq.~\ref{eq26}. In summary, we can include the multi-head self-attention into our unified formula as well.

\begin{equation}
    \begin{split}
    &\textbf{W}_{i,j,a,b,c,m} = Conv({\textbf{W}^{p}_{V}}_{:,:,c,m-p\times D_{h}},\sigma_{ab}(\textbf{A}^{p}_{i,j}))_{a,b} \\
    &\textbf{where}\ \ p = \lfloor m/D_{h} \rfloor
    \label{eq26}
    \end{split}
\end{equation}

\subsection{Relationship Between Self-attention and Convolution}

In this section, we will discuss the relationship between self-attention and convolution. To be specific, we will try to illustrate and analyze the approach to expressing convolution with self-attention under certain circumstance.

Paper~\cite{cordonnier2019relationship} proves that under certain sufficient conditions, a multi-head self-attention layer can simulate and express any convolutional layer. Their theorem can be illustrated as: A multi-head self-attention layer with $M$ heads of dimension $D_{h}$ and output dimension $D_{out}=M\times D_{h}$ can express any convolutional layer of kernel size $\sqrt{M}\times \sqrt{M}$ and $D_{h}$ output channels. The essence of the theorem is enforcing each attention head to focus specifically on the relationship between the central pixel and one pixel in certain position (the corresponding attention probability is 1), ignoring the connections with other pixels simultaneously (the corresponding attention probability is 0). Thus focusing on $M$ pixels of different positions at most, simulating the behaviour of one $\sqrt{M}\times \sqrt{M}$ convolution operator. As a matter of fact, this theorem is established when selecting certain relative positional encoding and supposing that the attention probabilities are independent of input $\textbf{X}$, thus setting $\textbf{W}_{K}=\textbf{W}_{Q}=0$. For $y_{ij}$ in pixel $(i,j)$ of the output, one special case of it can be obtained by Eq.~\ref{eq27-a}:

\begin{equation}
    \begin{split}
    \textbf{y}_{ij} &= \sum^{M-1}_{m=0} \left(\sum_a \sum_b \sigma(\textbf{A}^{m}_{i,j,l-a,l-b})\text{X}_{i-a,j-b,:}\right) \textbf{W}^{m} \\
    &= \sum^{M-1}_{m=0} \text{X}_{i-l+\lfloor m/l\rfloor,j-l+m-\lfloor m/l\rfloor\times l,:} \textbf{W}^{m} \\
    &\textbf{where} \ \ a,b \in [-l,l]\ ,\  l= \lfloor\frac{\sqrt{M}}{2}\rfloor \\
    &\textbf{A}^{m}_{i,j,l-a,l-b}=\left\{
\begin{aligned}
1 & , a=\lfloor m/l\rfloor-l \ , \ b=m-\lfloor m/l\rfloor\times l-l\\
0 & , otherwise
\end{aligned}
\right. \\
    &\textbf{W}^{m} = \textbf{W}_{V}^{m}{\textbf{W}_{O}}_{(m-1)\times D_{h}+1:m\times D_{h}+1,:}
    \label{eq27-a}
    \end{split}
\end{equation}

To put this in the unified formula, then the expression of the evolution kernel will be:

\begin{equation}
    \begin{split}
    \textbf{W}_{i,j,a,b,c,n}& = \textbf{W}^{a\times \sqrt{M} + b}_{c,n} \\
    &= \textbf{W}_{V}^{a\times \sqrt{M} + b}{\textbf{W}_{O}}_{(a\times \sqrt{M} + b-1)\times D_{h}+1:a\times \sqrt{M} + b\times D_{h}+1,:}
    \label{eq27-b}
    \end{split}
\end{equation}

However, we argue that this kind of equation waste most of the resource in the channel dimension, for that one attention head only focus on the connection between two pixels with certain relative distance while ignoring others. It actually reorganize the information of channels split by multiple heads to kernel spatial dimension. Let's examine and compare convolution and self-attention from the perspective of evolution, superficially, it is found that the difference lies in that the kernel of convolution is shared in each spatial position while the kernel for self-attention is generated specifically for each pixel. If we make the aggregation weights of self-attention kernel (both single-head and multi-head) remain the same for each pixel, in which case, self-attention is equivalent to convolution in the formula. That is:

\begin{subequations}
    \begin{align}
    &\textbf{W}_{i,j,a,b,c,m} = \textbf{W}_{p,q,a,b,c,m} \label{eq28-1} \\
    &Conv({\textbf{W}_{V}}_{:,:,c,m},\sigma_{ab}(\textbf{A}_{i,j}))_{a,b} = Conv({\textbf{W}_{V}}_{:,:,c,m},\sigma_{ab}(\textbf{A}_{p,q}))_{a,b}\label{eq28-2}\\
    &\textbf{A}_{i,j} = \textbf{A}_{p,q}\label{eq28-3}\\
    &\textbf{X}_{i,j,:}\textbf{W}_Q\textbf{W}^T_K\textbf{X}^T_{i-l+m,j-l+n,:} = \textbf{X}_{p,q,:}\textbf{W}_Q\textbf{W}^T_K\textbf{X}^T_{p-l+m,q-l+n,:}\label{eq28-4}
    \end{align}
\end{subequations}

To make the Eq.~\ref{eq28-4} establish, its value must be unaffected by the alternation of input $\textbf{X}$. As a result, $\textbf{W}_Q$ and $\textbf{W}_K$ must be set as zero tensors to meet the requirement. However, this approach makes the attention score map zero constantly, which is meaningless for later weight aggregation. Simulate the approach utilized in~\cite{vaswani2017attention}, we turn to the relative positional encoding. Then the attention score map $\textbf{A}^{rel}$ becomes:

\begin{equation}
    \begin{split}
    \textbf{A}^{rel}_{i,j,m,n} = \textbf{X}^T_{i,j,:}\textbf{W}^T_Q\textbf{W}_K\textbf{X}_{i-l+m,j-l+n,:} + \textbf{X}^T_{i,j,:}\textbf{W}^T_Q\widehat{\textbf{W}}_K \textbf{r}_{\delta} +
    \textbf{u}^T\textbf{W}_K\textbf{X}_{i-l+m,j-l+n,:} +
    \textbf{v}^T\widehat{\textbf{W}}_K \textbf{r}_{\delta}
    \label{eq28-a}
    \end{split}
\end{equation}

where $\textbf{u}$ and $\textbf{v}$ are two learned vectors, $\delta$ represents the shift between the query pixel $(i,j)$ and the key pixel $(i-l+m,j-l+n)$, and $\textbf{r}_{\delta} \in \mathbb{R}^{D_{p}}$ stands for the relative positional encoding. Moreover, ${\widehat{\textbf{W}}}_K$ is the key weights corresponding to the relative position of pixels. To make the attention map constant to different pixels and unaffected by the input $\textbf{X}$, it is only required to make $\textbf{W}_Q = \textbf{W}_K = \textbf{0}$. Thus, the final attention score map becomes:

\begin{equation}
    \begin{split}
    \textbf{A}^{rel}_{i,j} = \textbf{v}^T\widehat{\textbf{W}}_K \textbf{r}_{\delta}
    \label{eq28-b}
    \end{split}
\end{equation}

Thus the expression of kernel can be formulated as:

\begin{equation}
    \begin{split}
    \textbf{W}_{i,j,a,b,c,m} &= Conv({\textbf{W}_{V}}_{:,:,c,m},\sigma_{ab}(\textbf{A}^{rel}_{i,j}))_{a,b}\\
    &=Conv({\textbf{W}_{V}}_{:,:,c,m},\sigma_{ab}(\textbf{v}^T\widehat{\textbf{W}}_K \textbf{r}_{\delta})_{a,b}
    \label{eq28-c}
    \end{split}
\end{equation}

Superficially, Eq.~\ref{eq27-b} and Eq.~\ref{eq28-c} have the same expression as convolution kernel under the unified formula. However, it is not the case when we dig into the essence of the kernel generation. For Eq.~\ref{eq27-b}, it only considers the case of multi-head self-attention. The corresponding convolution kernel $\textbf{W}$, which should contains $M \times D_{in}\times D_{out}$ learned parameters, is derived from $\textbf{W}^{m}_{V}$ and $\textbf{W}_{O}$, which contain $M \times D_{in}\times D_{out}/M + D_{out}\times D_{out} $learned parameters in total. Clearly, the number of actual learned parameters is smaller than that required, thus the parameters obtained in $\textbf{W}$ are not updated individually and independently, in which case, $\textbf{W}$ here fails to make the same difference as the convolution kernel learned straightforwardly does. Actually, the kernel rank component $\textbf{W}^{m}$ is smaller than ${D_{out}/M}$ according to the principle of matrix multiply, hence it fails to be non-singular and can only express the convolution with ${D_{out}/M}$ output channels instead of $D_{out}$ output channels. This actually compensates spatial sources of kernel with channel sources. For Eq.~\ref{eq28-c}, the attention score is shared across both channels and positions, thus the actual convolution kernel is derived from one certain attention map. While it takes all channel information into consideration, it fails to discriminate the specific information contained in the space dimension of the kernel for that all pixels in one kernel slice are amplified or decayed the same times consistently. 
As a matter of fact, self-attention produces certain attention score map (can be seen as the predecessor of the kernel) in each position, however, this raw kernel with sufficient kernel space is shared across channels. Clearly, if we make the kernel both shared across the space on the basis of being shared across channels, there will be extremely little learned parameters and the expression ability of the model will be reduced significantly. Here, the effect of the value matrix that aggregation channel information can be ignored.

Thus we can conclude that self-attention utilizes the diversity of spatial information aggregation to compensate the monotonicity of channel information aggregation from the perspective of kernels. To make self-attention express the convolution, we have to compensate the channel source of the self-attention to the kernel space source, rather than making the kernel constant in each pixel simply.

\emph{One approach to address this problem is that we can generate $D_{in}$ attention score maps to make them specific to different channels (like the standard \emph{depth-wise} convolution) instead of one single map shared across channels in one pixel. This approach, which is both spatial- and channel-specific, is one special case of evolution.}

\subsection{Case of Involution}

In terms of the involution operator, suppose one involution kernel $\textbf{W}_{Inv} \in \mathbb{R}^{H\times W\times K\times K\times G}$, the parameter $G$ possesses the different essence compared to the \emph{group} in convolution and evolution. As a matter of fact, $G$ is similar to the multiple heads number $M$ in multi-head self-attention mechanism. Given $G = 1$, the values in the channel dimension of pixel $\textbf{X}_{(i,j)}$ share the same involution kernel. When $G > 1$, the n-th involution kernel in affects the values in the channel range of $[(n-1)\times {\lceil D_{in}/G \rceil},n\times {\lceil D_{in}/G \rceil}]$ in each position $(i,j)$, which bears the resemblance to the multiple attention score maps in the multi-head self-attention discussed above. Follow this thought, the involution kernel ${\textbf{W}_{Inv}}_{:,:,:,:,g}$ is similar to the $g-th$ attention score map, but both obtained according to different generation approaches. Besides, the later transformations performed at the input feature map are different as well. For the convenience of statement, we suppose $G=1$ in the following part.

Compared to the evolution kernel, involution kernel lacks two dimensions which stand for input channels and output channels. As is shown in Eq.~\ref{eq10}, the aggregation of involution operation is equivalent to the \emph{depth-wise} convolution (evolution) and makes no difference to the channel number of the feature map, which is meant to be input channel-agnostic. In this case, we can create the evolution kernel $\textbf{W}$ as expressed in Eq.~\ref{eq29} where the parameter $N = 1$. For the case of $G > 1$, the expression of evolution kernel is similar to Eq.~\ref{eq26}. Afterwards, we can obtain the evolution version of the involution operator according to Eq.~\ref{eq14} where $G$ here is set to be $D_{in}$.

\begin{equation}
    \begin{split}
    \textbf{W}_{i,j,a,b,1,m} = F(\textbf{X}) = {\textbf{W}_{Inv}}_{i,j,a,b,1} 
    \label{eq29}
    \end{split}
\end{equation}

Paper~\cite{li2021involution} claims that self-attention is actually a subclass of involution. However, we challenge the correctness and stringency of this argument. Firstly, in the case of involution, the number of output channels is the same with that of the input. While for self-attention, the number of output channels can be a user-specific hyper-parameter. Secondly, they ignore the projection performed on original input $\textbf{X}$ by $\textbf{W}_{V}$, which makes the input feature map inconsistency with that in Eq.~\ref{eq10}. Finally, they fail to take the circumstance of multi-head self-attention into consideration.

\subsection{Key Points of Evolution}

\textbf{Kernel Generation}
Kernel generation concerns about how to generate the kernel, namely, the form of Evolution Function. Generally, all the weights that utilize to extract features for in deep learning models can be generated by certain function. 
The Evolution Function can be input-agnostic, such as function for convolution with can be regarded as take the identity tensor as the input.
Also, the function can be input-specific, which means taking the feature map as the input and being related to the certain input, such as self-attention, involution, WeightNets~\cite{ma2020weightnet} and other dynamic convolutions~\cite{yang2019condconv,zhang2020dynet,chen2020dynamic,li2022omni}.
Obviously, these two kind of functions can be integrated as they are instantiations for Evolution.

\textbf{Aggregation Scope}
For Evolution Kernel, the aggregation scope represents the certain scope for interacting the information among the pixels and extracting the features, which can be regarded as the receptive field.
While for Evolution Function, the aggregation scope stands for the set of input pixels being taken consideration for the generation of Evolution Kernels.

\textbf{Channel and Space Preferences}
Similar to the aggregation scope, we can talk about the preferences of channel and space from the perspective of kernels, as well as the kernel generation functions.
There are four combinations of different channel and space preferences, namely spatial- and channel-agnostic, spatial- and channel-specific, spatial-agnostic and channel-specific, as well as spatial-specific and channel-agnostic.
\emph{specific} means the weights are shared across the certain dimension, while the weights are unrelated to each other in terms of \emph{agnostic}.  
Classically, the kernel generation function is spatial-agnostic and channel-specific (spatial- and channel-agnostic for the case of convolution). Whether the essence of the kernel generation function be inverted or consistent? The answer is positive when we view it from a even higher perspective that there may exist one meta function to generate the current kernel generation function. This can be nested within multiple layers and be multi-order, hence it is complex and maybe meaningless. This paper only discuss the one-order kernel generation functions, for the multi-order kernel generation will be leaved as the future work.

\section{Conclusion}
This paper raises Evolution, a unified formula for different feature operators from a high-level perspective, which utilize Evolution Function to generate Evolution Kernels, which extract and aggregate the features from the feature maps.
To prove the unification, we equivalently transform the existing mathematical formula of three typical operators, namely, convolution, self-attention and involution, and then deduce their formulas into the framework of Evolution in detail.
In addition, we discuss different forms of the Evolution Function and analyse the properties of Evolution Kernels correspondingly generated by different Evolution Functions.
We hope Evolution can give inspirations to the further research and innovations of powerful deep learning feature operators.

%\bibliographystyle{unsrt}  
%\bibliography{references}  

\end{document}